\pdfoutput=1

\documentclass[11pt]{article}

\usepackage{acl}

\usepackage{bm} 
\usepackage{times}
\usepackage{latexsym}

\usepackage[T1]{fontenc}

\usepackage[utf8]{inputenc}

\usepackage{amsmath}
\usepackage{microtype}
\usepackage{newfloat}
\usepackage{listings}
\usepackage{lipsum}
\usepackage{graphicx}
\usepackage{enumitem}
\usepackage{xspace}
\usepackage{xcolor}
\usepackage{booktabs}
\usepackage{multirow}
\usepackage{array}
\usepackage{newfloat}
\usepackage{listings}

\newcommand{\method}{\texttt{MindMap}\xspace} 
\newcommand{\bmG}{\mathcal{G}}
\newcommand{\bmV}{\mathcal{V}}

\title{\method: Knowledge Graph Prompting Sparks Graph of Thoughts in Large Language Models}

\author{Yilin Wen$^1$ \\
  \texttt{yilinwen510@gmail.com} \\
  Zifeng Wang$^1$  \\
  University of Illinois Urbana-Champaign, Champaign, IL\\
  \texttt{zifengw2@illinois.edu} \\ 
  Jimeng Sun$^2$ \\
  University of Illinois Urbana-Champaign, Champaign, IL\\
  \texttt{jimeng@illinois.edu} \\}

\begin{document}
\maketitle
\begin{abstract}

Large language models (LLMs) have achieved remarkable performance in natural language understanding and generation tasks. However, they often suffer from limitations such as difficulty in incorporating new knowledge, generating hallucinations, and explaining their reasoning process. To address these challenges, we propose a novel prompting pipeline, named \method, that leverages knowledge graphs (KGs) to enhance LLMs' inference and transparency. Our method enables LLMs to comprehend KG inputs and infer with a combination of implicit and external knowledge. Moreover, our method elicits the mind map of LLMs, which reveals their reasoning pathways based on the ontology of knowledge. We evaluate our method on diverse question \& answering tasks, especially in medical domains, and show significant improvements over baselines. We also introduce a new hallucination evaluation benchmark and analyze the effects of different components of our method. Our results demonstrate the effectiveness and robustness of our method in merging knowledge from LLMs and KGs for combined inference. To reproduce our results and extend the framework further, we make our codebase available at \href{https://github.com/wyl-willing/MindMap}{https://github.com/wyl-willing/MindMap}.
\end{abstract}

\section{Introduction}

\begin{figure}[t]
    \centering
    \includegraphics[width=\linewidth]{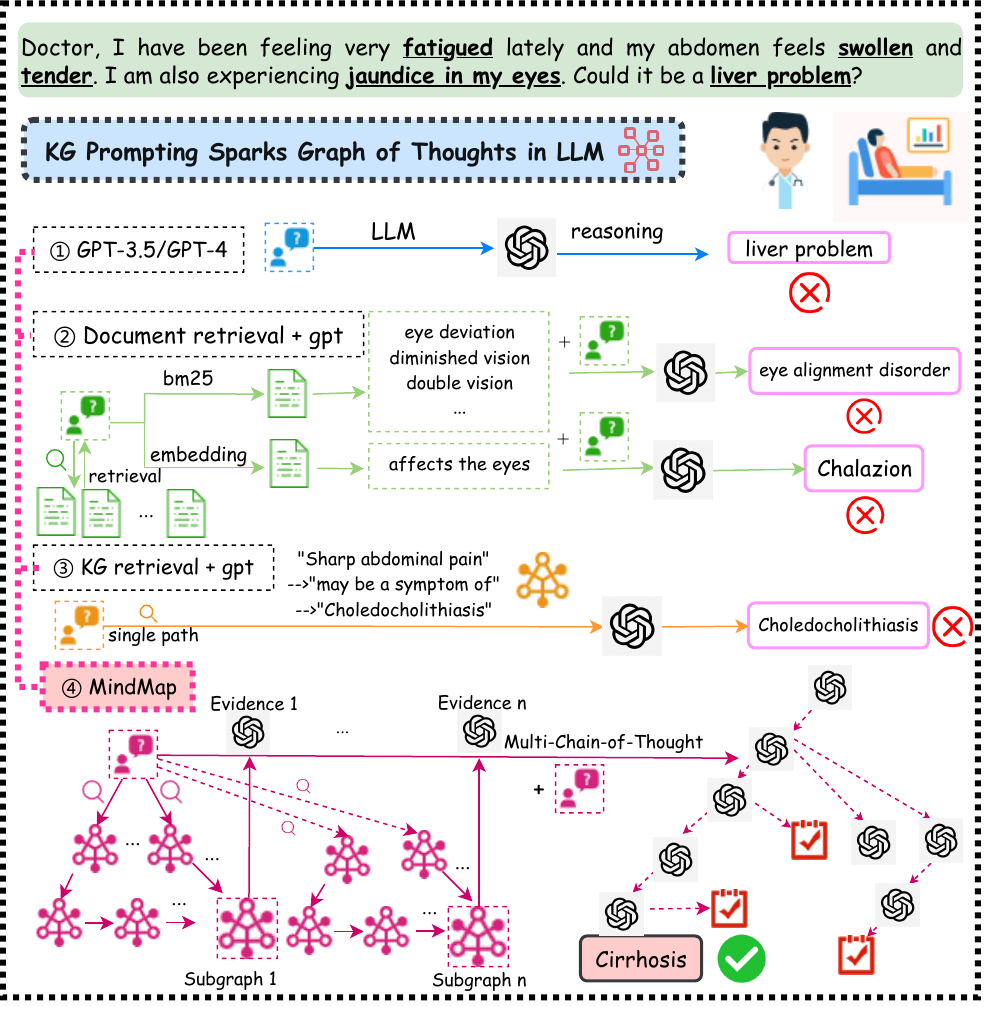}
    \caption{A conceptual comparison between our method and the other prompting baselines: LLM-only, document retrieval + LLM, and KG retrieval + LLM.}
    \label{fig:demo}
    \vspace{-1em}
\end{figure}

Scaling large language models (LLMs) to billions of parameters and a training corpus of trillion words was proved to induce surprising performance in various tasks \cite{brown2020language,chowdhery2022palm}. Pre-trained LLMs can be adapted to domain tasks with further fine-tuning \cite{singhal2023large} or be aligned with human preferences with instruction-tuning \cite{ouyang2022training}. Nonetheless, several hurdles lie in the front of steering LLMs in production:

\begin{figure*}[!t]
    \centering
    \includegraphics[width=\linewidth]{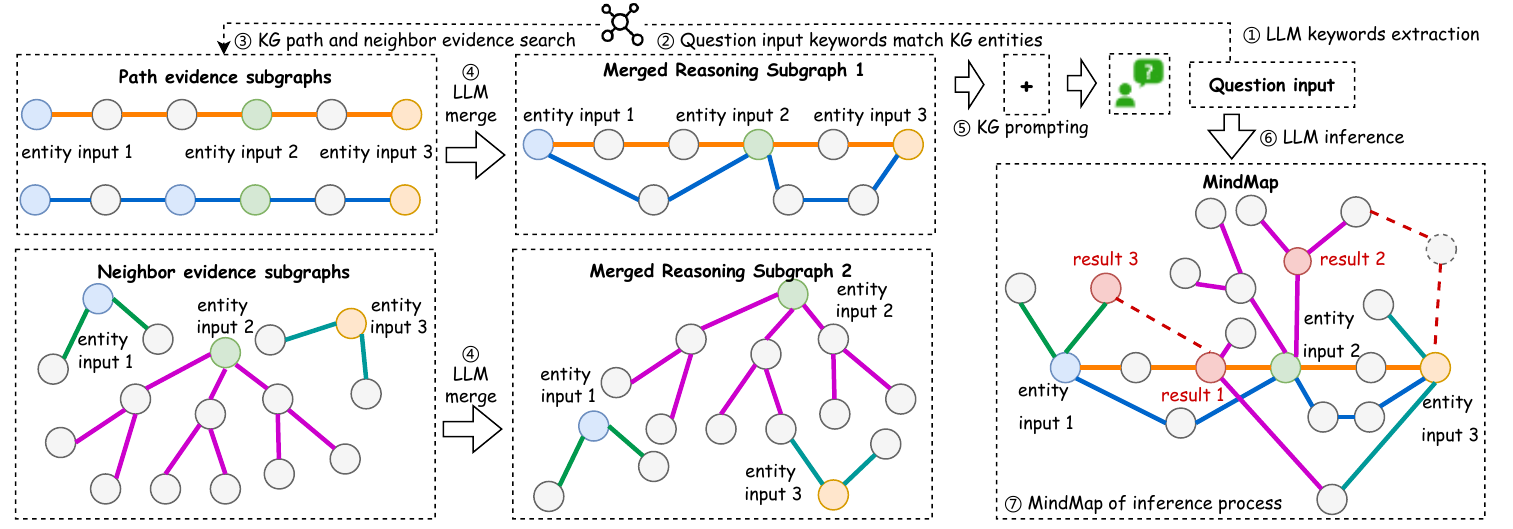}
    \caption{A conceptual demonstration of evidence query sub-graphs, merged reasoning sub-graphs, and mind map. The entity inputs $\mathcal{V}_q$ is identified from the input. Lines and circles of the same color indicate that they correspond. The red dashed lines in the \method box illustrate the augmentation operation based on the knowledge of LLM.\label{fig:concept_graph} }
    \vspace{-1em}
\end{figure*}

\begin{itemize}[leftmargin=*]
    \item \textbf{Inflexibility}. The pre-trained LLMs possess outdated knowledge and are inflexible to parameter updating. Fine-tuning LLMs can be tricky because either collecting high-quality instruction data and building the training pipeline can be costly \cite{cao2023instruction}, or continually fine-tuning LLMs renders a risk of catastrophic forgetting \cite{razdaibiedina2022progressive}.
    \item \textbf{Hallucination}. LLMs are notoriously known to produce hallucinations with plausible-sounding but wrong outputs \cite{ji2023survey}, which causes serious concerns for high-stake applications such as medical diagnosis.
    \item \textbf{Transparency}. LLMs are also criticized for their lack of transparency due to the black-box nature \cite{danilevsky2020survey}. The knowledge is implicitly stored in LLM's parameters, thus infeasible to be validated. Also, the inference process in deep neural networks remains elusive to be interpretable.
\end{itemize}

As a classic way to build large-scale structural knowledge bases, knowledge graphs (KG) are established by the triples of entities and relations, i.e., $\{\texttt{head}, \texttt{relation}, \texttt{tail}\}$. They can provide explicit knowledge representation and interpretable reasoning paths. Besides, KGs are amenable to continual modifications to debug the existing knowledge or add new knowledge. Due to their flexibility, preciseness, and interpretability, KGs emerged as a promising complement to the drawbacks of LLMs \cite{llm_kg}. For instance, KG triples were added to the training of LLMs \cite{zhang2019ernie} or KG encoders were entangled with LLM layers for joint inference and optimization on graph and text data \cite{zhang2021greaselm}. By contrast, our work pivots on the synergistic inference of KGs and fixed LLMs, which is applicable to powerful pre-trained LLMs, such as commercial LLM-as-service APIs. In general, the prior arts in this venue can be categorized into two genres:

\begin{itemize}[leftmargin=*]
    \item  \textbf{Retrieval-Augmented LLM Inference}. Researchers tried to retrieve documents to augment LLM inference \cite{lewis2020retrieval} while suffering from inaccurate retrieval and lengthy documents \citep{liu2023lost}. Recently, several attempts were made to incorporate extracted KG triples into the prompt to LLMs to answer KG-related questions \cite{baek2023knowledge}. However, this approach treats KG inputs as plain text and ignores their graphical structure, which causes the generated response to be hard to validate and vulnerable to hallucinations.
    \item \textbf{Graph Mining with LLMs}. There were also attempts to prompt LLMs to comprehend graphical inputs, while they primarily experimented with graph mining tasks, e.g., edge detection and graph summarization \cite{guo2023gpt4graph,chen2023exploring}. It was rarely explored in text generation tasks that require complex reasoning across multiple evidence graphs grounded on KGs.
\end{itemize}

The goal of this work is to build a plug-and-play prompting approach to elicit the graph-of-thoughts reasoning capability in LLMs. We call our method \method because it enables LLMs to comprehend graphical inputs to build their own mind map that supports evidence-grounded generation.  A conceptual demonstration of our framework is in Figure~\ref{fig:concept_graph}. Specifically, \method sparks the graph of thoughts of LLMs that (1) consolidates the retrieved facts from KGs and the implicit knowledge from LLMs, (2) discovers new patterns in input KGs, and (3) reasons over the mind map to yield final outputs. We conducted experiments on three datasets to illustrate that \method outperforms a series of prompting approaches by a large margin. This work underscores how LLM can learn to conduct synergistic inference with KG. By integrating both implicit and explicit knowledge, LLMs can achieve transparent and dependable inference, adapting to different levels of correctness in additional KG information.


\section{Related Work}
\noindent \textbf{Prompt Engineering}. The ``pre-train, prompt, and predict" paradigm has become the best practice for natural language processing in few-shot or zero-shot manners \cite{liu2023pre}. The core insight is LLMs are able to adapt to new tasks following the input context and instructions via in-context learning \cite{brown2020language}, especially with instruction tuning \cite{wei2022finetuned} and alignment \cite{ouyang2022training}. Retrieval-augmented generation emerged as a way to dynamically inject additional evidence for LLM inference \citep{lewis2020retrieval}. The common practice is to query a dense embedding database to find the relevant document pieces to the input user questions, then put the retrieved corpus back to the prompt input. However, documents can be lengthy, thus not fitting into the context length limit of LLM. It was also identified even though we can build long documents as prompts, LLMs usually fail to capture information in the middle of the prompt and produce hallucinations \citep{liu2023lost}. Another line of research looks to prompt to elicit the intermediate reasoning steps of LLMs in chains \citep{wei2023chainofthought} and trees \citep{yao2023tree}, while these approaches all focus on eliciting the implicit knowledge from LLMs. Nonetheless, our work explores sparking the reasoning of LLMs on graph inputs, with an emphasis on joint reasoning with implicit and external explicit knowledge.

\noindent \textbf{Knowledge Graph Augmented LLM}. Researchers have explored using knowledge graphs (KGs) to enhance LLMs in two main directions: (1) integrating KGs into LLM pre-training and (2) injecting KGs into LLM inference. For (1), it is a common practice to design knowledge-aware training objectives by either putting KG entities and relations into the training data \cite{zhang2019ernie,sun2021ernie} or applying KG prediction tasks, e.g., link prediction, as additional supervision \cite{yasunaga2022deep}. However, when scaling the pre-training data to a web-scale corpus with trillion words, it is intractable to find or create KGs with approximate scale. More importantly, although these methods directly compress KG knowledge into LLM's parameters via supervision, they do not mitigate the fundamental limits of LLMs in flexibility, reliability, and transparency. 

For (2), the early efforts were centered around fusing KG triples into the inputs of LLMs via attention \cite{liu2020k,sun2020colake} or attaching graph encoders to LLM encoders to process KG inputs \cite{wang2019improving}. The follow-ups further adopted graph neural networks in parallel to LLMs for joint reasoning \cite{yasunaga2021qa} and added interactions between text tokens and KG entities in the intermediate layers of LLMs \cite{zhang2021greaselm,yao2023beyond}. Witnessing the recent success of pre-trained LLMs, the research paradigm is shifting to prompting fixed pre-trained LLMs with graphical inputs. This line of research includes prompting LLMs for KG entity linking prediction \cite{choudhary2023complex,sun2023think}, graph mining \cite{guo2023gpt4graph}, and KG question answering \cite{baek2023knowledge}. While these approaches permit LLMs to comprehend graph inputs, they either target KG tasks exclusively or recall retrieved facts and translate them to plain texts, ignoring the structure of KG. Most importantly, these methods often rely heavily on the factual correctness of the KG and ignore the situation where the KG does not match the question.

\section{Method}
\begin{figure*}[t]
    \centering
    \includegraphics[width=\linewidth]{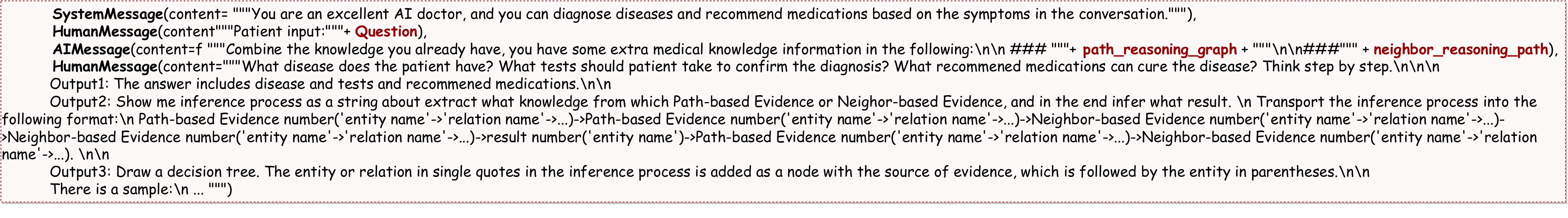}
    \caption{The prompt template for final input to LLM. Its input is the question and reasoning graphs.}
    \label{fig:prompt}
    \vspace{-1em}
\end{figure*}

We show the framework of \method in Figure~\ref{fig:method}, which comprises three main components:
\begin{enumerate}[leftmargin=*]
    \item \textbf{Evidence graph mining}: We begin by identifying the set of entities  $\mathcal{V}_q$ from the raw input and query the source KG $\bmG$ to build multiple \textit{evidence sub-graphs} $\mathcal{G}_q$.
    \item \textbf{Evidence graph aggregation}: Next, LLMs are prompted to comprehend and aggregate the retrieved evidence sub-graphs to build the \textit{reasoning graphs} $\mathcal{G}_m$.
    \item \textbf{LLM reasoning on the mind map}: Last, we prompt LLMs to consolidate the built reasoning graph and their implicit knowledge to generate the answer and build a \textit{mind map} explaining the reasoning process.
\end{enumerate}

\subsection{Step I: Evidence Graph Mining}
Discovering the relevant evidence sub-graphs $\bmG_q$ from the external KG breaks down into two main stages.


\subsubsection{Entity Recognition} 
We first use LLM to identify key entities from the question query $Q$. Specifically, we use a prompt that consists of three parts: the question to be analyzed, the template phrase "The extra entities are", and two examples. The full prompt is given in Table \ref{tab:ner} of Appendix \ref{appx:prompt}. We then apply BERT similarity to match entities and keywords. Specifically, we encode all the keyword entities $M$ extracted by LLM and all the entities $\bmG$ from the external knowledge graph into dense embeddings $H_{M}$ and $H_{\bmG}$ respectively, and then compute the cosine similarity matrix between them. For each keyword, we obtain the entity set $\bmV_q$ with the highest similarity scores, which we use to build the evidence subgraph in the next step.


\subsubsection{Evidence Sub-graphs Exploration} 

We define the extra source knowledge graph by $\mathcal{G}  = \left\{ {\left\langle {u,r,o} \right\rangle \left| {u \in \psi,r \in \varphi,o \in \mathcal{L} } \right.} \right\}$, where $\psi_q$, $\varphi_q$, and $\mathcal{L}_q$ represent the entity set, relation set, and textual set, respectively. The objective of this stage is to build the evidence sub-graphs $\bmG_q=\{\bmG_q^{path},\bmG_q^{nei}\}$ based on the extracted entities $\bmV_q$. An evidence sub-graph is defined by  ${\mathcal{G}_q^*} = \left( {{\mathcal{N}_q^*},{\mathcal{E}_q}^*},{\psi^*_q},{\varphi^*_q},{\mathcal{L}^*_q}\right)$, where $\mathcal{N}_q^*$ is the node set, ${\mathcal{E}_q^*}$ is the edge set where each edge $e = \left\langle {n,n'} \right\rangle, \ n,n^\prime \in \mathcal{N}_q^*$.

As shown in Figure \ref{fig:method}, we use two approaches to build the evidence sub-graph set $\mathcal{G}_q$ from the source knowledge graph. \textbf{(1)} Path-based exploration traces intermediary paths within $\bmG$ to connect important entities from the query. We form path segments by exploring connected nodes from a chosen node in $\mathcal{V}_q^0$ for at most $k$ hops. The process continues until all segments are connected, creating a set of sub-graphs stored in $\bmG_{q}^{\text{path}}$. \textbf{(2)} Neighbor-based exploration adds related knowledge by expanding each node $n$ in $\mathcal{N}_q$ by 1-hop to its neighbors, adding triples $\{(n,e,n^\prime)\}$ to $\bmG_q^{\text{nei}}$. This approach incorporates additional query-related evidence into $\mathcal{G}_q$. After exploration, we update $\mathcal{V}_q$ with newly added intermediate nodes from bridging pathways. To manage information overhead and maintain diversity, we prune $\bmG_q^{\text{path}}$ and $\bmG_q^{\text{nei}}$ by clustering and sampling sub-graphs based on their head entities. These pruning steps result in the final evidence graph $\bmG_q$, optimizing information while preserving diversity. Specific details are shown in Appendix E. We show the hallucination influence of results using path-based exploration and neighbor-based exploration components in the experiment part.

\subsection{Step II: Evidence Graph Aggregation}

In this phase, LLM is instructed to consolidate the diverse evidence sub-graphs $\mathcal{G}_q^*$ into a unified \textit{reasoning graph} $\bmG_m$. This reasoning graph $\bmG_m$, upon completion, serves as an external augmented graph input for Step III, providing a holistic perspective on all evidence sub-graphs to enhance the output generation process.

To generate the final additional knowledge subgraph input, we first extracted at least $k$ path-based and $k$ neighbor-based evidence subgraphs from the previous part, each representing a possible connection between the query entities. Then, we formatted each subgraph as an entity chain, such as \textit{ ``(Fatigue, Nausea) - IsSymptomOf - LiverProblem"}, and assigned a sequence number, such as $P_1$, $P_2$, $N_1$ , $N_2$. Next, we prompted LLM to convert each entity chain into a natural language description, using a template that can be found in Table \ref{tab:nlp} of Appendix \ref{appx:prompt}, and defined them as reasoning graph $\bmG_m$. This design had two benefits: \textbf{(a)} It simplified the subgraphs into a concise and consistent format that captured the essential information. \textbf{(b)} It leveraged LLM's natural language understanding and generation abilities to unify semantically similar entities and resolve potential ambiguities.

\subsection{Step III: LLM Reasoning with Mind Map}
In this step, LLMs are prompted with two reasoning graphs $\bmG_m^{\text{path}}$ and $\bmG_m^{\text{nei}}$ in Step II to produce the final outputs.

\subsubsection{Prompting for Graph Reasoning}
To generate a mind map and find final results, we provide LLMs with a prompt that has five components: a system instruction, a question, evidence graphs $\bmG_m$, a graph-of-thought instruction, and exemplars. The graph-of-thought instruction uses the Langchain technique to guide LLMs to comprehend and enhance the input, build their own mind map for reasoning, and index the knowledge sources of the mind map. The prompt template is detailed in Figure \ref{fig:prompt}. The final answers consist of a summary answer, an inference process, and a mind map that shows the graph reasoning pathways. The entities in the mind map are from the evidence graphs $\bmG_m$ and LLM's own retrieval enhancement, as shown in the right red box in Figure~\ref{fig:method} in Appendix \ref{appx:prompt}.


\subsubsection{Synergistic Inference with LLM and KG Knowledge}
We find that previous retrieval-augmented LLMs tend to rephrase the retrieved facts without exploiting the knowledge of LLM itself. However, \method enables LLM to synergistically infer from both the retrieved evidence graphs and its own knowledge. We attribute this ability to three aspects: \textbf{(1)} \textit{Language Understanding}, as LLM can comprehend and extract the knowledge from $\bmG_m$ and the query in natural language, \textbf{(2)} \textit{Knowledge Reasoning}, as LLM can perform entity disambiguation and produce the final answer based on the mind map constructed from $\bmG_m$, and \textbf{(3)} \textit{Knowledge Enhancement}, as LLM can leverage its implicit knowledge to expand, connect, and improve the information relevant to the query. This ability is especially valuable when the external knowledge input is inaccurate.

\section{Experiments}
\begin{table}[t]
\centering
\caption{The statistics of the used datasets. \label{tab:data_stats}}
\setlength{\extrarowheight}{2pt}
\resizebox{1\linewidth}{!}{
\begin{tabular}{cccc}
\specialrule{2pt}{0pt}{0pt}
\textbf{Dataset} & \textbf{GenMedGPT-5k}       & \textbf{CMCQA}       & \textbf{ExplainCPE}  \\ \hline
Domain     & English Clinical Q\&A & Chinese Long Dialogue      & 5-way Choice Question  \\
Multi-task          & Disease, Drug, Test & Disease, Drug, Test, Food &  Option, Explanation     \\
KG dataset  & EMCKG & CMCKG       & \textbf{CMCKG}    \\ \hline
Question     & 714  & 468       & 400                          \\ 
Node & 1122 & 62282      & 62282 \\
Triple  & 5802 & 506490     & 506490 \\
Relationship & 6 & 12     & 12\\
\specialrule{2pt}{0pt}{0pt}
\end{tabular}}
\end{table}


\begin{table}[t]
\centering
\caption{The BERTScore and GPT4 ranking of all methods for \textbf{GenMedGPT-5k}.}\label{tab:genmedgpt_rank}
\setlength{\extrarowheight}{2pt}
\resizebox{1\linewidth}{!}{
\begin{tabular}{lccccc}
\specialrule{2pt}{0pt}{0pt}
\multirow{2}{*}{}  & \multicolumn{3}{c}{\textbf{BERT Score}}          & \textbf{GPT4 Ranking} & \textbf{Hallucination}\\
                   & \textbf{Precision}   & \textbf{Recall}      & \textbf{F1 Score}    & \textbf{(Average)}          &   Quantify   \\ \specialrule{2pt}{0pt}{0pt}
\method            & \textbf{0.7936} & 0.7977  & \textbf{0.7954} & \textbf{1.8725}      & \textbf{0.6070}    \\ \specialrule{2pt}{0pt}{0pt}
GPT-3.5         & 0.7612 & 0.8003 & 0.7800  & 4.8571       & 0.5563      \\
Tree-of-thought(TOT) & 0.7202 & 0.7949 & 0.7554 & - & 0.5483      \\
GPT4               & 0.7689 & 0.7893 & 0.7786 & 4.1764     & 0.5577        \\ \hline
BM25 Retriever      & 0.7693 & 0.7981 & 0.7831 & 3.5546      & 0.5834       \\
Embedding Retriever& 0.7690 & 0.8038 & 0.7857  & 3.1232      & 0.5886         \\
KG Retriever  & 0.7717 & 0.8030 & 0.7868 & 3.4159        & 0.5871     \\ 
\specialrule{2pt}{0pt}{0pt}
\end{tabular}}
\end{table}

\begin{table*}[!ht]
\centering
\caption{The pair-wise comparison by GPT-4 on the winning rate of \method v.s. baselines on diversity \& integrity score (\%), fact total match score (\%), and disease diagnosis (\%), on \textbf{GenMedGPT-5k}.} \label{tab:win_rate_genmedgpt}
\setlength{\extrarowheight}{2pt}
 \resizebox{\textwidth}{!}{
\begin{tabular}{lcccccccccccccccccc}
\specialrule{3pt}{0pt}{0pt}
\method vs Baseline                    & \multicolumn{3}{c}{\textbf{GPT-3.5}}            & \multicolumn{3}{c}{\textbf{BM25 Retriever}}      & \multicolumn{3}{c}{\textbf{Embedding Retriver}} & \multicolumn{3}{c}{\textbf{KG Reriever}}        & \multicolumn{3}{c}{\textbf{GPT-4}}      &   \multicolumn{3}{c}{\textbf{TOT}}       \\
Metries  & \textbf{Win}     & \textbf{Tie} & \textbf{Lose} & \textbf{Win}      & \textbf{Tie} & \textbf{Lose} & \textbf{Win}     & \textbf{Tie} & \textbf{Lose} & \textbf{Win}     & \textbf{Tie} & \textbf{Lose} & \textbf{Win}     & \textbf{Tie} & \textbf{Lose} & \textbf{Win}   & \textbf{Tie} & \textbf{Lose} \\ \specialrule{2pt}{0pt}{0pt}
Diversity \& integrity   & 100            & -            & -             & 100             & -            & -             & 100            & -            & -             & 100            & -            & -             & 100            & -            & -           & - & - & -  \\
Total factualness           & \textbf{80.11} & -            & 19.89       & 66.67           & -            & 33.33       & 76.05          & -            & 23.95       & 73.53          & -            & 26.47       & 75.77          & -            & 24.23     &  78.5 & - & 21.5  \\
Disease diagnosis           & 84.73          & 0.14       & 15.13       & 75.91           & 1.26       & 22.83       & 77.03          & 1.96       & 21.01       & 66.67          & 2.94       & 30.39       & 73.11          & 1.40       & 25.49   & 75 & 24.6 & 0.3   \\
Drug recommendation          & 88             & 5          & 7           & 87              & 8          & 5           & 72             & 13         & 15          & 74             & 19         & 7           & 83             & 8          & 9        & 87 & 5 & 8    \\ \hline
Average                                & \textbf{88.21} & 1.285      & 10.505      & \textbf{82.395} & 2.315      & 15.29       & \textbf{81.27} & 3.74       & 14.99       & \textbf{78.55} & 5.485      & 15.965      & \textbf{82.97} & 2.35       & 14.68    & \textbf{80.17} & 14.8 & 9.93   \\ \specialrule{2pt}{0pt}{0pt}
\end{tabular}}
\end{table*}

\begin{table}[!ht]
\centering
\caption{The BERTScore and GPT-4 ranking of all methods for \textbf{CMCQA} dataset.}\label{tab:cmcqa_rank}
\setlength{\extrarowheight}{2pt}
\resizebox{1\linewidth}{!}{
\begin{tabular}{lcccc}
\specialrule{2pt}{0pt}{0pt}
\multirow{2}{*}{}  & \multicolumn{3}{c}{\textbf{BERT Score}}          & \textbf{GPT-4 Ranking} \\
                   & \textbf{Precision}   & \textbf{Recall}      & \textbf{F1 Score}    & \textbf{(Average)}               \\ \specialrule{2pt}{0pt}{0pt}
\method            & \textbf{0.9415} & 0.9321 & 0.9367 & \textbf{2.3}     \\ \specialrule{2pt}{0pt}{0pt}
GPT-3.5         & 0.9385 & 0.9361  & 0.9372 & 3.4     \\
GPT-4              & 0.9355 & 0.9358 & 0.9356 & 3.6      \\ \hline
BM25 Retriever      & 0.9365 & 0.9348 & 0.9356 & 3.7     \\
Embedding Retriever & 0.9357 & 0.9334 & 0.9345 & 5.4     \\\hline
KG Retriever  & 0.9318 & 0.9348 & 0.9332 & 2.3     \\ \specialrule{2pt}{0pt}{0pt}
\end{tabular}}
\end{table}

\begin{table*}[!ht]
\centering
\caption{The pair-wise comparison by GPT-4 on the winning rate of \method v.s. baselines on disease diagnosis and drug recommendation on CMCQA.}
\setlength{\extrarowheight}{2pt}
 \resizebox{\textwidth}{!}{
\begin{tabular}{lccccccccccccccc}
\specialrule{2pt}{0pt}{0pt}
\method vs Baseline & \multicolumn{3}{c}{\textbf{GPT-3.5}}        & \multicolumn{3}{c}{\textbf{BM25 Retriever}} & \multicolumn{3}{c}{\textbf{Embedding Retriver}} & \multicolumn{3}{c}{\textbf{KG Reriever}}    & \multicolumn{3}{c}{\textbf{GPT-4}}           \\
Metrics             & \textbf{Win} & \textbf{Tie} & \textbf{Lose} & \textbf{Win} & \textbf{Tie} & \textbf{Lose} & \textbf{Win}   & \textbf{Tie}  & \textbf{Lose}  & \textbf{Win} & \textbf{Tie} & \textbf{Lose} & \textbf{Win} & \textbf{Tie} & \textbf{Lose} \\ \specialrule{2pt}{0pt}{0pt}
Disease diagnosis   & 35.68      & 39.96      & 24.36       & 30.98      & 50.21      & 18.80       & 37.18        & 42.74       & 20.08        & 34.40      & 45.51      & 20.09       & 27.99      & 47.22      & 24.79       \\
Drug recommendation & 47.32      & 30.62      & 22.06       & 47.11      & 29.34      & 23.55       & 44.97        & 32.12       & 22.91        & 44.33      & 31.26      & 24.41       & 44.11      & 29.76      & 26.12       \\ \hline
Average             & 41.5       & 35.29      & 23.21       & 39.045     & 39.775     & 21.175      & 41.075       & 37.43       & 21.495       & 39.365     & 38.385     & 22.25       & 36.05      & 38.49      & 25.455      \\ \specialrule{2pt}{0pt}{0pt}
\end{tabular}}
\label{tab:win_rate_cmcqa}
\end{table*}
\begin{table}[t]
\centering
\caption{The accuracy scores for \textbf{ExplainCPE}. We calculate the rates of correct, wrong, and failed responses.}\label{tab:acc_explaincpe}
\setlength{\extrarowheight}{2pt}
\resizebox{0.8\linewidth}{!}{
\begin{tabular}{lcccc}
\specialrule{2pt}{0pt}{0pt}
  \textbf{Method}   & \multicolumn{3}{c}{\textbf{Accuracy Rate(\%)}}      \\
\textbf{}           & \textbf{Correct} & \textbf{Wrong} & \textbf{Failed} \\ \specialrule{2pt}{0pt}{0pt}
GPT-3.5             & 52.2             & 47.2           & 0.5             \\
BM25 Retriever      & 50               & 44.2           & 5.7             \\
Embedding Retriever & 54.2             & 45.2           & 0.5             \\
KG Retriever        & 42               & 44             & 14              \\
GPT-4                & 72               & 27.7           & 0.2             \\
 \specialrule{2pt}{0pt}{0pt}
\method      & \textbf{61.7}    & \textbf{37.7}  & \textbf{0.5}    \\
w/o prompt template $\mathbf{p}_1$  & 53.5            & 46             & 0.5             \\
\specialrule{2pt}{0pt}{0pt}
\end{tabular}}
\end{table}

\begin{table}[t]
\centering
\caption{Quantitative comparison with BERTScore and GPT-4 preference ranking between \method and baselines in ExplainCPE dataset.}\label{tab:explaincpe_gpt4_explanation}
\setlength{\extrarowheight}{2pt}
\resizebox{1\linewidth}{!}{
\begin{tabular}{lcccc}
\specialrule{2pt}{0pt}{0pt}
\multirow{2}{*}{}  & \multicolumn{3}{c}{\textbf{BERT Score}}          & \textbf{GPT-4 Ranking} \\
                   & \textbf{Precision}   & \textbf{Recall}      & \textbf{F1 Score}    & \textbf{(Average)}               \\ \specialrule{2pt}{0pt}{0pt}
\method            & 0.9335            & 0.9376         & 0.9354           & \textbf{2.98}     \\\specialrule{2pt}{0pt}{0pt}
GPT-3.5         & 0.9449            & 0.9487         & 0.9467           & 3.0425                                     \\
GPT-4               & 0.9487            & 0.9529         & 0.9507           & 3.0075    \\ \hline
BM25 Retriever      & 0.9413            & 0.9411         & 0.9411           & 3.6675                                     \\
Embedding Retriever & 0.9440            & 0.9459        & 0.9449           & 4.3175                                     \\
KG Retriever  & 0.9354             & 0.9373         & 0.9362           & 3.985           \\
\specialrule{2pt}{0pt}{0pt}
\end{tabular}}
\end{table}  

\begin{table}[t]
\centering
\caption{The BERTScore and hallucination qualification of different component for GenMedGPT-5k.}\label{tab:ablation}
\setlength{\extrarowheight}{2pt}
\resizebox{1\linewidth}{!}{
\begin{tabular}{lccccc}
\specialrule{2pt}{0pt}{0pt}
\multirow{2}{*}{} &  \textbf{Tokens}  & \multicolumn{3}{c}{\textbf{BERT Score}}          &  \textbf{Hallucination} \\
        &   (Average)            & \textbf{Precision}   & \textbf{Recall}      & \textbf{F1 Score}       &   Quantify   \\ \specialrule{2pt}{0pt}{0pt}
Path-only    & 1028    & 0.6310 & 0.7885 & 0.7002  & 0.3854             \\
Neighbor-only & 1236 & 0.6393 & 0.7930 & 0.7072 & 0.3894  \\\specialrule{2pt}{0pt}{0pt}
\method      & 1431      & \textbf{0.7938} & \textbf{0.7987}  & \textbf{0.7960} & \textbf{0.5890}       \\ \specialrule{2pt}{0pt}{0pt}
Improved-path & +403 & +0.1628 & +0.0102 & +0.0957 & +0.2036 \\
Improved-neigh & +195 & +0.1545 & +0.0057 & +0.0888 & +0.1996 \\
\specialrule{2pt}{0pt}{0pt}
\end{tabular}}
\end{table}

We evaluate our method for a suite of question \& answering tasks that require sophisticated reasoning and domain knowledge and compare it with retrieval-based baselines.

\subsection{Experimental Setup}

We evaluate the utilization of external knowledge graphs by \method in complex question-answering tasks across three medical Q$\&$A datasets: \textit{GenMedGPT-5k}, \textit{CMCQA}, and \textit{ExplainCPE}. These datasets cover patient-doctor dialogues, multi-round clinical dialogues, and multiple-choice questions from the Chinese National Licensed Pharmacist Examination, respectively. To support KG-enhanced methods, we construct two knowledge graphs (\textit{EMCKG} and \textit{CMCKG}) containing entities and relationships related to medical concepts. The \textit{ExplainCPE} dataset utilizes \textit{CMCKG} with knowledge mismatches to assess the impact of incorrect retrieval knowledge on model performance. We compare \method's ability to integrate implicit and explicit knowledge with various baselines, including vanilla GPT-3.5 and GPT-4, as well as the tree-of-thought method (TOT) \cite{yao2023tree}, which uses a tree structure for reasoning. Additionally, we consider three retrieval-augmented baselines: \textit{BM25 retriever}, \textit{Text Embedding retriever}, and \textit{KG retriever}, see instruction details in Appendix \ref{appx:prompt}. These baselines leverage different methods and sources for evidence retrieval, with gpt-3.5-turbo-0613 as the backbone for all retrieval-based methods. Detailed descriptions of these baselines are provided in Appendix \ref{appx:baseline}.

\subsection{Medical Question Answering}
We used GenMedGPT-5K to test how LLMs deal with question-answering in the medical domain, where LLMs need to answer with disease diagnosis, drug recommendation, and test recommendation.

\subsubsection{Evaluation Metrics}
We used two metrics, \textit{BERTScore}\citep{zhang2019bertscore} and \textit{GPT-4 Rating}, for quantitative evaluation. \textit{BERTScore} measures semantic similarity between the generated and reference answers. GPT-4 was employed to (1) rank answer quality against ground truth and (2) compare pairs of answers on four criteria: \textit{response diversity and integrity}, \textit{overall factual correctness}, \textit{correctness of disease diagnosis}, and \textit{correctness of drug recommendation}. In addition, we introduce a new metric for hallucination quantification, which estimates the degree of deviation from the facts in the generated answers \cite{liang2023uhgeval}. To compute this metric, we first use the question-extra entities data generated by Step I and train a keyword extraction model (NER-MT5) based on mT5-large. Then, we input the outputs of \method, other baselines, and labels into the NER-MT5 model to obtain the lists of keywords for each answer. Finally, we concatenate the keywords with commas as ner-sentences, and calculate the tfidf similarity score between the ner-sentences of different outputs. A lower score indicates more hallucination in the answer.

\subsubsection{Results}
In Table \ref{tab:genmedgpt_rank}, various methods are evaluated based on BERTScore, GPT-4 ranking scores, and hallucination quantification scores. While BERTScore shows similar results among methods, \method exhibits a slight improvement, possibly due to the shared tone in medical responses. However, for medical questions, comprehensive domain knowledge is crucial, not well-captured by BERTScore. GPT-4 ranking scores and hallucination quantification reveal that \method significantly outperforms others, with an average GPT-4 ranking of 1.8725 and low hallucination scores. This underscores \method's ability to generate evidence-grounded, plausible, and accurate answers compared to baseline models like GPT-3.5 and GPT-4, which may produce incorrect responses due to reliance on implicit knowledge. Additionally, Table~\ref{tab:win_rate_genmedgpt} demonstrates \method's consistent superiority over other methods, emphasizing the value of integrating external knowledge to mitigate LLM hallucinations and provide accurate answers.

\subsection{Long Dialogue Question Answering}
In our experiments on the CMCQA dataset, characterized by lengthy dialogues requiring complex reasoning, Table~\ref{tab:cmcqa_rank} showcases \method consistently ranking favorably compared to most baselines, albeit similar to KG Retriever. Additionally, in Table~\ref{tab:win_rate_cmcqa}, \method consistently outperforms baselines in pairwise winning rates as judged by GPT-4. Despite a narrower performance gap compared to GenMedGPT-5K, attributed to the inadequacy of the knowledge graph (KG) in covering all necessary facts for CMCQA questions, \method still outshines all retrieval-based methods, including KG Retriever. This suggests previous retrieval-based approaches might overly rely on retrieved external knowledge, compromising the language model's (LLM) ability to grasp intricate logic and dialogue nuances using its implicit knowledge. Conversely, \method leverages both external and implicit knowledge in graph reasoning, yielding more accurate answers.

\subsection{Generate with Mismatch Knowledge from KG}
\begin{figure*}[!htbp]
    \centering
    \includegraphics[width=\linewidth]{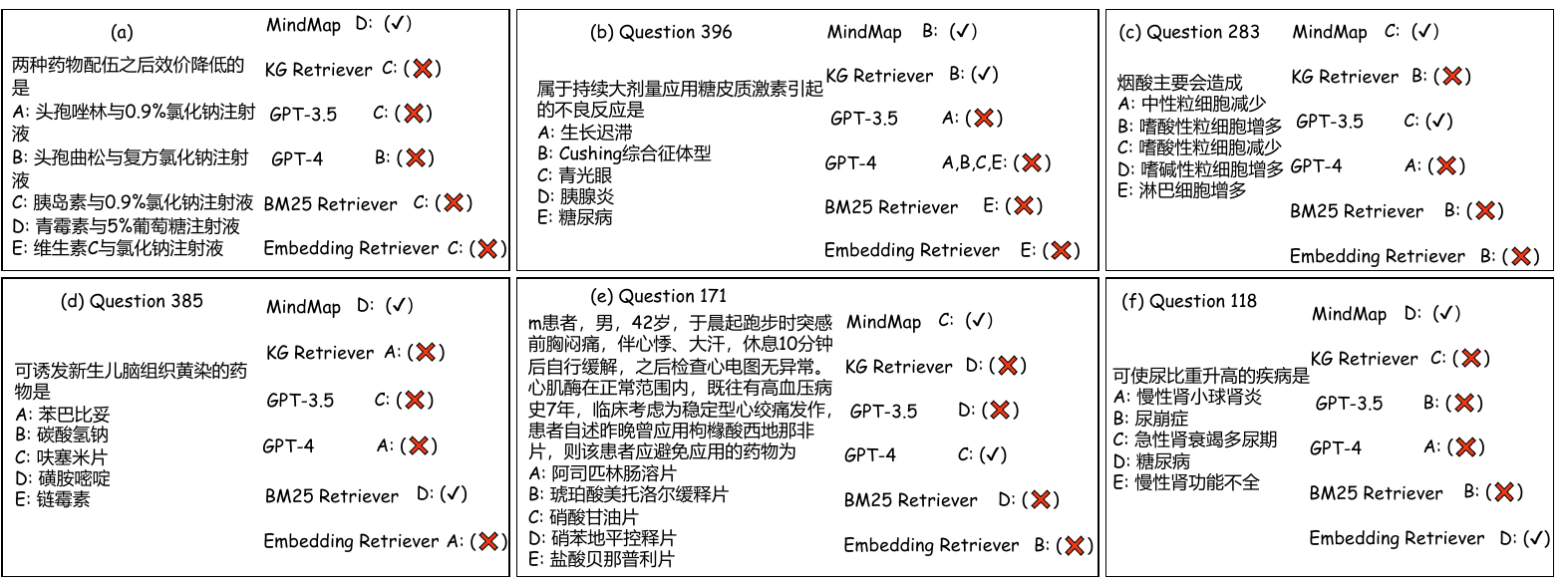}
    \caption{\textbf{Case examples of multi-choice in ExplainCPE}, comparing predictions by Baselines and \method.} 
   \label{fig:case4}
\end{figure*}
In addressing the robustness of \method concerning the factual correctness of KG, we leverage the identical KG dataset employed in the second dataset - \textit{ExplainPE}. Consequently, the knowledge retrieved may tend to be redundant or devoid of accurate information. This aspect is particularly crucial since it mirrors a common scenario in production, where LLM often needs to generate answers by amalgamating both its implicit knowledge and the knowledge retrieved from external sources.


\subsubsection{Evaluation Metrics}
We evaluate all methods based on the accuracy of the generated choice and the quality of the explanations. For assessing explanation quality, we use BERTScore and GPT-4 ranking. We specifically instruct the GPT-4 rater to prioritize the correctness of the explanation over its helpfulness or integrity.


\subsubsection{Results}
In Table \ref{tab:acc_explaincpe}, our method (\method) demonstrates superior accuracy compared to various baselines, affirming its effectiveness over document retrieval prompting techniques. Interestingly, we observed that directly incorporating retrieved knowledge into prompts sometimes degrades answer quality, as seen with KG Retriever and BM25 Retriever performing worse than the vanilla GPT-3.5 model. This discrepancy arises from mismatched external knowledge, leading to misleading effects on the language model (LLM). The model tends to rely on retrieved knowledge, and when inaccurate, the LLM may generate errors. Notably, GPT-4 outperforms GPT-3.5+\method, possibly due to test questions being part of GPT-4's pre-training corpus. Ablation analysis on instruction prompts revealed that prompting the LLM to "combine with the knowledge you already have" ($\mathbf{p}_1$) improved performance by 8.2$\%$. Moreover, Table \ref{tab:explaincpe_gpt4_explanation} highlights \method's ability to generate rationales for answers, earning a ranking of 2.98 by GPT-4.


\subsection{Ablation Study}
   
In our study, we compared our method (\method) with two variants: Neighbor-only and Path-only. Neighbor-only focuses on neighbor-based evidence exploration, while Path-only concentrates on path-based evidence exploration. Despite using additional tokens, \method showed significant improvements in hallucination quantification compared to both Neighbor-only and Path-only methods. This highlights the importance of combining both path-based and neighbor-based approaches to reduce hallucinations. Notably, the neighbor-based method proved more effective in enhancing factual accuracy compared to the path-based method. For tasks involving medical inquiries, path-based methods are better at finding relevant external information, though they struggle with multi-hop answers such as medication and test recommendations.

\subsection{In-depth Analysis}

We further conducted an in-depth analysis of the cases by \method, focusing on the discussion of the following aspects.

\subsubsection{How does \method perform without correct KG knowledge?}
In Figure~\ref{fig:case4}(c) (Appendix \ref{appx:case}), when faced with a question where GPT-3.5 is accurate but KG Retriever errs, \method achieves an accuracy rate of 55\%. We attribute the low accuracy of the KG Retriever to its inability to retrieve the necessary knowledge for problem-solving. \method effectively addresses such instances by leveraging the LLM inherent knowledge, identifying pertinent external explicit knowledge, and seamlessly integrating it into a unified graph structure.

\subsubsection{How robust is \method to unmatched fact queries?}
The question in Figure~\ref{fig:case1} (Appendix \ref{appx:case}) contains misleading symptom facts, such as \textit{`jaundice in my eyes'} leading baseline models to retrieve irrelevant knowledge linked to \textit{`eye'}. This results in failure to identify the correct disease, with recommended drugs and tests unrelated to liver disease. In contrast, our model \method accurately identifies \textit{cirrhosis'} and recommends the relevant \textit{`blood test'} showcasing its robustness.

\subsubsection{How does \method aggregate evidence graphs considering entity semantics?}
In Figure \ref{fig:case2} of Appendix \ref{appx:case}, nodes like \textit{`vaginitis'} and \textit{`atrophic vaginitis'} are present in different evidence sub-graphs but share a semantic identity. \method allows LLMs to disambiguate and merge these diverse evidence graphs for more effective reasoning. The resulting mind maps also map entities back to the input evidence graphs. Additionally, Figure \ref{fig:case2} illustrates the GPT-4 rater's preference for total factual correctness and disease diagnosis factual correctness across methods. Notably, \method is highlighted for providing more specific disease diagnosis results compared to the baseline, which offers vague mentions and lacks treatment options. In terms of disease diagnosis factual correctness, the GPT-4 rater observes that \method aligns better with the ground truth.

\subsubsection{How does \method visualize the inference process and evidence sources?}
Figure \ref{fig:case3} in Appendix \ref{appx:case} presents a comprehensive response to a CMCQA question. It includes a summary, an inference process, and a mind map. The summary extracts the accurate result from the mind map, while the inference process displays multiple reasoning chains from the entities on the evidence graph $\bmG_m$. The mind map combines all the inference chains into a reasoning graph, providing an intuitive understanding of knowledge connections in each step and the sources of evidence sub-graphs.

\subsubsection{How does \method leverage LLM knowledge for various tasks? }
Figure \ref{fig:case4} in Appendix \ref{appx:case} illustrates \method's performance on diverse question types. For drug-related questions (a) and (d), which demand in-depth knowledge, \method outperforms other methods. Disease-related questions (b) and (f) show comparable results between retrieval methods and \method, indicating that incorporating external knowledge mitigates errors in language model outputs. Notably, for general knowledge questions (c), LLMs like GPT-3.5 perform better, while retrieval methods lag. This suggests that retrieval methods may overlook the knowledge embedded in LLMs. Conversely, \method performs as well as GPT-3.5 in handling general knowledge questions, highlighting its effectiveness in synergizing LLM and KG knowledge for adaptable inference across datasets with varying KG fact accuracies.

\section{Conclusion}
This paper introduced knowledge graph (KG) prompting that 1) endows LLMs with the capability of comprehending KG inputs and 2) facilitates LLMs inferring with a combined implicit knowledge and the retrieved external knowledge. We then investigate eliciting the mind map, where LLMs perform the reasoning and generate the answers with rationales represented in graphs.  Through extensive experiments on three question $\&$ answering datasets, we demonstrated that our approach, \method, achieves remarkable empirical gains over vanilla LLMs and retrieval-augmented generation methods, and is robust to mismatched retrieval knowledge. We envision this work opens the door to fulfilling reliable and transparent LLM inference in production.


\bibliography{main}
\clearpage
\appendix

\setcounter{secnumdepth}{2} 

\begin{appendix}
\section{Construction of Datasets}\label{appx:datasets}

 \begin{figure*}[t]
    \centering
    \includegraphics[width=\linewidth]{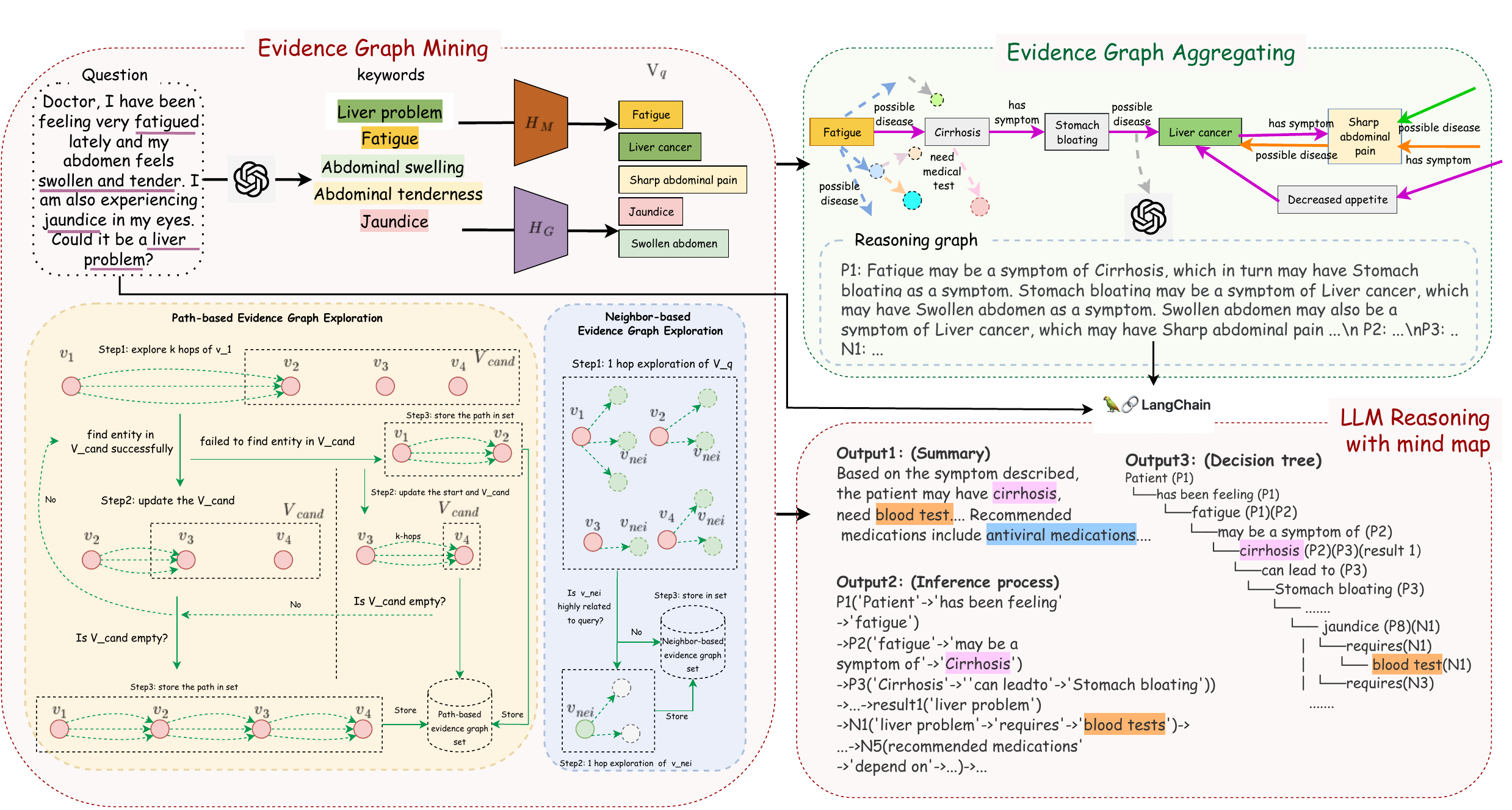}
    \caption{An overview of the architecture of our proposed \method. The left part illustrates the components of \textit{evidence graph mining}, while the right part shows the \textit{evidence graph aggregation} and \textit{LLM reasoning with mind map}.}
    \label{fig:method}
    \vspace{-1em}
\end{figure*}

\begin{itemize}[leftmargin=*]
    \item \textbf{GenMedGPT-5k} is a 5K generated dialogue between patients and GPT-3.5 grounded on a disease database\footnote{\url{https://github.com/Kent0n-Li/ChatDoctor}}. The question describes the symptoms of the patient during the consultation, which comes from the iCliniq database. Based on the database, the generated answers cover the diagnosis, symptoms, recommended treatments, and medical tests. We sampled 714 dialogues as the test set.
    \item \textbf{CMCQA} contains multi-round dialogues between patients and doctors in Chinese. It covers materials from 45 clinical departments such as andrology, gynecology, and obstetrics and gynecology. We simplified the setup by merging the patient's questions and the clinician's answers to build the one-round Q\&A. We sampled 468 from all to build the test set.
    \item \textbf{ExplainCPE} is a 5-way choice question dataset from the Chinese National Licensed Pharmacist Examination. Answering the questions requires a series of capabilities, including logical reasoning, drug knowledge, scenario analysis, mathematical calculation, disease knowledge, and general knowledge. The answers include the correct options and the explanations. We extracted 400 samples related to disease diagnosis and treatment recommendations from the original dataset for testing.

\end{itemize}

\section{Implementation of Knowledge Graph}\label{appx:KG}
\begin{itemize}[leftmargin=*]
    \item \textbf{EMCKG} We utilized a disease database\footnote{\url{https://github.com/Kent0n-Li/ChatDoctor/blob/main/format_dataset.csv}} to build the KG from scratch to support the knowledge source for the inference on GenMedGPT-5k. This database encompasses a diverse set of diseases and the corresponding symptoms, medical tests, treatments, etc. The entities in the EMCKG include disease, symptom, drug recommendation, and test recommendation. The relationships in the EMCKG include \textit{`possible$\_$disease'}, \textit{`need$\_$medical$\_$test'}, \textit{`need$\_$medication'}, \textit{`has$\_$symptom'}, \textit{`can$\_$check$\_$disease'}, \textit{`possible$\_$cure$\_$disease'}. In total, the yielded KG contains of 1122 nodes and 5802 triples.
\item \textbf{CMCKG} We established a KG based on \textit{QASystemOnMedicalKG}\footnote{\url{https://github.com/liuhuanyong/QASystemOnMedicalKG/blob/master/data/medical.json}} to support KG-augmented inference on CMCQA and ExplainCPE. The CMCKG includes various entities such as disease, symptom, syndrome, recommendation drugs, recommendation tests, recommendation food, and forbidden food. The relationships in the CMCKG include \textit{`has$\_$symptom'}, \textit{`possible$\_$disease'}, \textit{`need$\_$medical$\_$test'}, \textit{`has$\_$syndrome'}, \textit{`need$\_$recipe'}, \textit{`possible$\_$cure$\_$disease'}, \textit{`recipe$\_$\_is$\_$good$\_$for$\_$disease'}, \textit{`food$\_$\_is$\_$good$\_$for$\_$disease'}, \textit{`food$\_$\_is$\_$bad$\_$for$\_$disease'}, \textit{`need$\_$medication'}, \textit{`need$\_$food'}, and \textit{`forbid$\_$food'}. In total, the KG contains 62282 nodes, 12 relationships, and 506490 triples.

\end{itemize}

\section{Implementation of Baselines}\label{appx:baseline}
\begin{itemize}[leftmargin=*]
    \item \textbf{GPT-3.5 \& GPT-4} We evaluate the performance of the recent dominant LLM models as two baselines, using \textit{gpt-3.5-turbo} \cite{wang2023evaluating, ateia2023ChatGPT} and \textit{gpt-4}\footnote{\url{https://openai.com/gpt-4}} \cite{ali2022performance, guo2023gpt4graph} API respectively. 
\item \textbf{BM25 document retriever + GPT-3.5} We compare with existing BM25 document retriever methods \cite{roberts2020much, peng2023check}, which use BM25 retrieval scores \cite{robertson2009probabilistic} as logits when calculating $p(z|x)$. For fair comparisons, we use the same KG database as our method to generate different document files. Specifically, we use the GPT-3.5 API to convert all knowledge data centered on one disease into natural language text as the content of a document. For GenMedGPT-5k, we make 99 documents based on English medical KG $\mathcal{G}_{English}$. For CMCQA and ExplainCPE, we make 8808 documents based on Chinese medical KG $\mathcal{G}_{Chinese}$. For each question query, we retrieve the top $k$ gold document contexts based on bm25 scores.
\item \textbf{Text embedding document retrieval + GPT-3.5} Same as BM25 document retriever methods, text embedding document retrieval methods \cite{sharma2023ontology,lewis2020retrieval} retrieve the top $k$ documents for each question query. The difference is that in this method we train a word2vec embedding \cite{dai2020word2vec} on the document corpus as the evidence source for document ranking.
\item \textbf{KG retrieval + GPT-3.5} We compare with existing KG retrieval methods \cite{jia2021complex, sun2023think}, which aim to find the shortest KG path between every pair of question entities. The final prompt is then retrieved from KG to guide GPT-3.5 model to answer questions. For fair comparisons, we use the same preliminary process as our method to recognize the entities in question query. The key differences between \method and these are that they do not think on multiple evidence KG sub-graphs with multi-thought in LLM, and without backtracking evidence sources.
\item \textbf{Tree-of-thought (TOT)} We compare TOT as a typical Chain-of-thought \cite{NEURIPS2022_9d560961} baseline with \method. TOT is a method that uses a tree structure to solve complex problems\cite{yao2023tree}. By extending one inference path into multiple inference paths, the model can synthesize the results of multiple inference paths to obtain the final conclusion.

\end{itemize}

\section{Prompt Engine}\label{appx:prompt}
\begin{itemize}
\item \textbf{The instructions of \method components.} Table \ref{tab:ner} shows the instruction of Step I: entity recognition, which aims to identify and label medical entities in the user query. Table \ref{tab:nlp} shows the templates of Step II (Evidence Graph Aggregation), which generates natural language sentences from the evidence graph nodes and edges.
\item \textbf{The instructions of baseline methods:} Table \ref{tab:document} shows the prompt template of two document retrieval methods (BM25 Retrieval and Embedding Retrieval). The input is the question and the most related document context.
\item \textbf{The instructions of evaluation:} Figure \ref{fig:prompt} presents the final prompt used by \method for generating results and constructing a mind map. The prompt consists of a system message acknowledging the AI's expertise as a doctor, a user message representing the patient's input, and an AI message incorporating knowledge obtained from an external KG. The Langchain technique is employed to create the prompt, which guides the generation of step-by-step solutions. The response consists of a summary answer to the query, the inference process, and a mind map. Table \ref{tab:gpt4} illustrates an example of the pairwise ranking evaluation using the GPT-4 rater, which compares the quality of different responses based on various criteria.
\end{itemize}

\section{Evidence Subgraphs Exploration}\label{appx:path}
We provide more details on the path-based and neighbor-based exploration methods as follows:
\begin{itemize}
\item \textbf{Path-based Evidence Graph set $\bmG_{q}^{\text{path}}$ Exploration} connects entities in $\mathcal{V}_q$ by tracing their intermediary pathways within $\bmG$: \textbf{(a)} Choose one node in $\mathcal{N}_q^0$ as the start node $n_1$. Place the remaining nodes in a candidate node set $\mathcal{N}_{cand}$. Explore at most $k$ hops from $n_1$ to find the next node $n_2$, where $n_2 \in \mathcal{N}_{cand}$. If $n_2$ is successfully reached within $k$ hops, update the start node as $n_2$ and remove $n_2$ from $\mathcal{N}_{cand}$. If $n_2$ cannot be found within $k$ hops, connect the segments of paths obtained so far and store them in $\bmG_{q}^{\text{path}}$. Then choose another node ${n_1}'$ from $\mathcal{N}_{cand}$ as the new start node, and remove both $n_1$ and $n_2$ from $\mathcal{N}_{cand}$. \textbf{(b)} Check if $\mathcal{N}_{cand}$ is empty. If it is not empty, iterate step 1 to find the next segment of the path. If it is empty, connect all segments to build a set of sub-graphs and put them into $\bmG_{q}^{\text{path}}$.
\item \textbf{Neighbor-based Evidence Graph set $\bmG_q$ Exploration} aims to incorporate more query-related evidence into $\bm{G}_q$. It has two steps: \textbf{(a)} Expand for each node $n\in \mathcal{V}_q$ by 1-hop to their neighbors $\{n\prime\}$ to add triples $\{(n,e,n^\prime)\}$ to $\bmG_q^{\text{nei}}$. \textbf{(b)} For each $v^\prime$, check if it is semantically related to the question. If so, further expand the 1-hop neighbors of $n^\prime$, adding triples $\left( {{n_{nei}},e',n'} \right)$ to $\bmG_q^{\text{nei}}$.
\end{itemize}

\section{In-depth Analysis}\label{appx:case}
\begin{figure*}[!thbp]
    \centering
    \includegraphics[width=\linewidth]{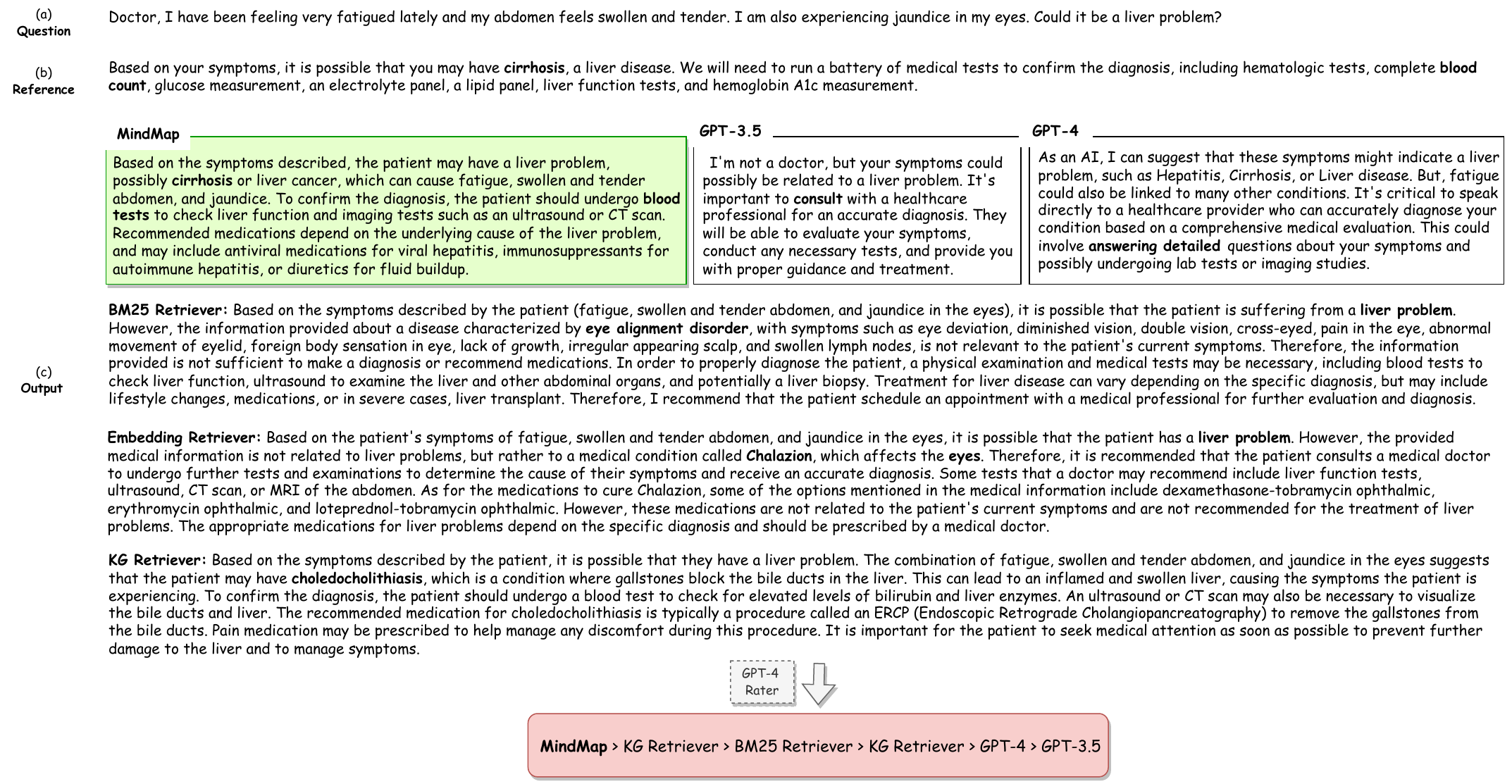}
    \caption{\textbf{A case compares \method and baselines with mismatched retrieved knowledge}, evaluated by the GPT factual correctness preference rater.} 
   \label{fig:case1}
\end{figure*}

\begin{figure*}[!thbp]
    \centering
    \includegraphics[width=\linewidth]{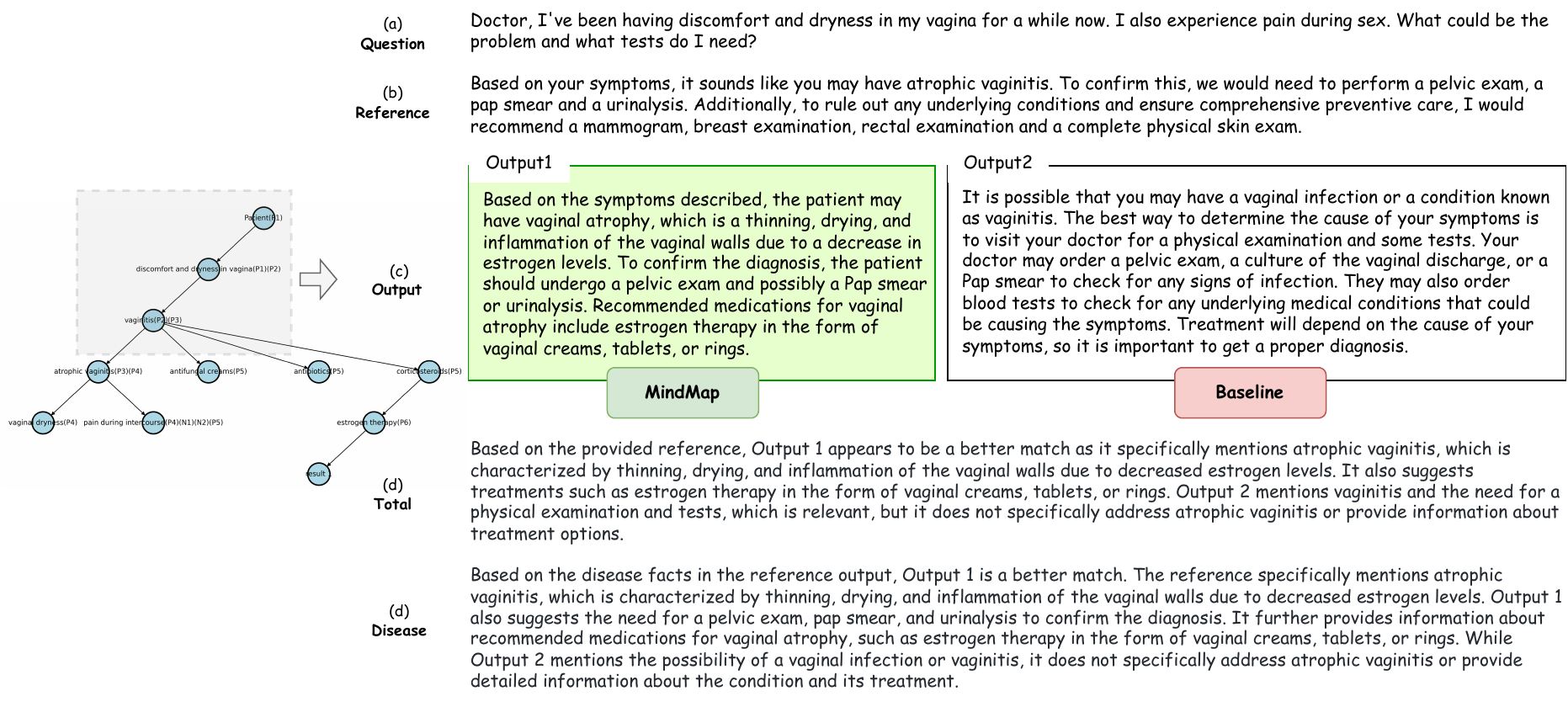}
    \caption{\textbf{Factually correctness evaluation in GenMedGPT-5k using GPT-4 preference ranking}: \method shows a  strong ability in fact-matching subtasks of question-answering by generating a mind map.} 
   \label{fig:case2}
\end{figure*}

\begin{figure*}[!thbp]
    \centering
    \includegraphics[width=\linewidth]{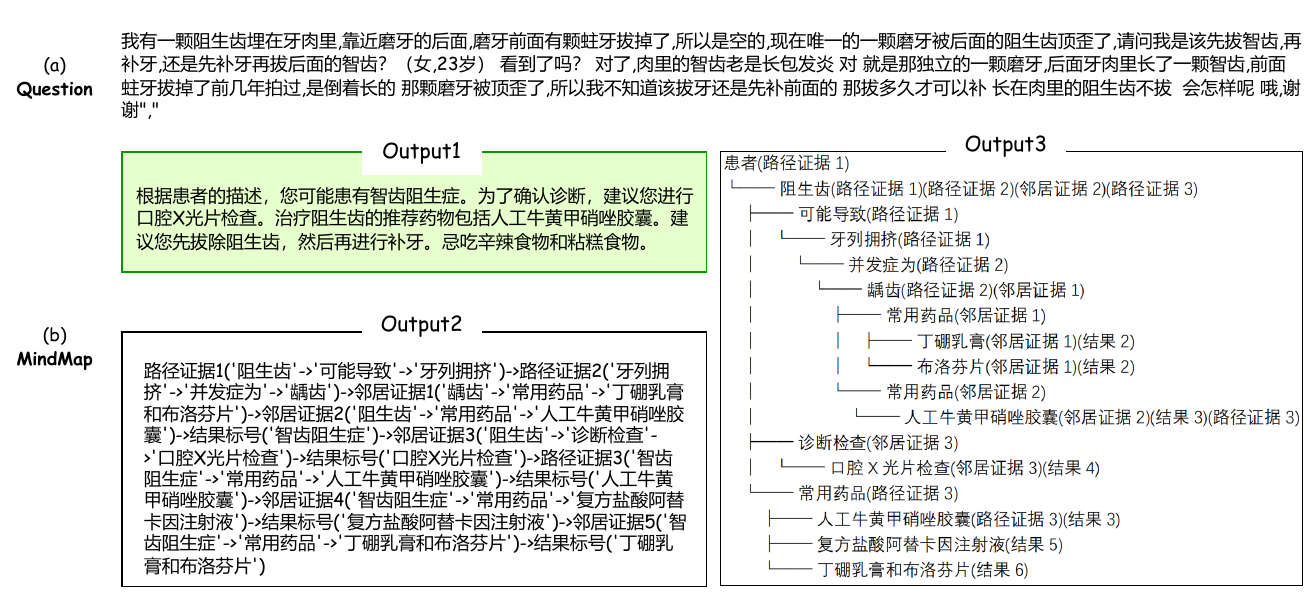}
    \caption{\textbf{An example to show the visualization of \method}. By generating mind maps, \method guides LLM to obtain the correct factual outputs for different subtasks.} 
   \label{fig:case3}
\end{figure*}

We select four examples for in-depth analysis, as shown in Figure \ref{fig:case1}, \ref{fig:case2}, \ref{fig:case3}, and \ref{fig:case4}. 
\begin{itemize}
    \item Figure \ref{fig:case1} presents an example from GenMedGPT-5k. It includes the question, reference response, the response generated by \method, responses from baselines, and the factual correctness preference determined by the GPT-4 rater. This example is used to discuss the robustness of \method in handling mismatched facts.

    \item Figure \ref{fig:case2} illustrates another example from GenMedGPT-5k. It displays the question query, reference response, summary responses from both \method and baseline models, a mind map generated by \method, and specific preferences in terms of factual correctness and sub-task disease fact match determined by the GPT-4 rater. This example shows the ability of \method to aggregate evidence graphs.

    \item Figure \ref{fig:case3} showcases an example from CMCQA. It includes the question query, a summary answer, the inference process, and the generated mind map by \method. This example provides insights into the visualization of the final output produced by \method.

    \item Figure \ref{fig:case4} demonstrates an example from ExplainCPE. It consists of six questions categorized into three different question types and evaluates the accuracy of \method and baseline models. This example allows us to examine the performance of \method across various tasks.
\end{itemize}

\section{Pairwise Ranking Evaluation}
For each pair of answers, as an example in Table \ref{tab:gpt4}, raters were asked to select the preferred response or indicate a tie along the following axes (with exact instruction text in quotes):
\begin{itemize}
    \item \textbf{Diversity and integrity:} ``According to the result in reference output, which output is better."
    \item \textbf{Total factual correctness:} ``According to the facts of disease diagnosis and drug and tests recommendation in reference output, which output is better match."
    \item \textbf{Disease diagnosis:} ``According to the disease diagnosis result in reference output, which output is better match."
    \item \textbf{Drug recommendation:} ``According to the drug recommendation result in reference output, which output is better match."
    
\end{itemize}
Note that for the second dataset CMCQA, since the reference label is derived from the actual dialogue answer, it may not contain facts. When the GPT-4 rater performs pairwise ranking evaluation, it is very easy to judge it as a tie. Therefore, we add an additional instruction: ``\textit{If they are the same, output "2". Try to output "1" or "0"}'', so as to force the rater to make a preference judgment.

\section{Limitations and Potential Risks}
The integration of knowledge graphs (KGs) with large language models (LLMs), particularly in medical contexts, presents several potential challenges. One significant concern is the risk of replicating any existing biases or errors in the knowledge graphs. These graphs, often built from pre-existing data sources, might contain outdated or partial information, which could inadvertently influence the LLM's outputs. Another issue lies in the integration complexity between KGs and LLMs, which could lead to unexpected errors or logical inconsistencies, especially when addressing intricate or vague queries. This aspect is critically important in the medical field, where precision is paramount. Moreover, there's a possibility that the LLMs might become excessively dependent on the KGs, which could hinder their performance in scenarios where KGs are not accessible or are lacking in information. Additionally, the use of "mind maps" to trace the LLMs’ reasoning paths, while innovative, raises questions about the models' interpretability. If these visual representations are complex or obscure, it may be difficult for users to understand how conclusions were reached, potentially diminishing trust in these advanced systems. In summary, while the merger of KGs with LLMs is a promising development, it is crucial to address these potential issues to ensure the responsible and efficacious application of this technology.

\begin{table*}[!thbp]
\centering
\begin{lstlisting}[language=Python, basicstyle = \ttfamily, breaklines = true]
template = """
There are some samples:
\n\n
### Instruction:\n'Learn to extract entities from the following medical questions.'\n\n### Input:\n
<CLS>Doctor, I have been having discomfort and dryness in my vagina for a while now. I also experience pain during sex. What could be the problem and what tests do I need?<SEP>The extracted entities are\n\n ### Output:
<CLS>Doctor, I have been having discomfort and dryness in my vagina for a while now. I also experience pain during sex. What could be the problem and what tests do I need?<SEP>The extracted entities are Vaginal pain, Vaginal dryness, Pain during intercourse<EOS>
\n\n
Instruction:\n'Learn to extract entities from the following medical answers.'\n\n### Input:\n
<CLS>Okay, based on your symptoms, we need to perform some diagnostic procedures to confirm the diagnosis. We may need to do a CAT scan of your head and an Influenzavirus antibody assay to rule out any other conditions. Additionally, we may need to evaluate you further and consider other respiratory therapy or physical therapy exercises to help you feel better.<SEP>The extracted entities are\n\n ### Output:
<CLS>Okay, based on your symptoms, we need to perform some diagnostic procedures to confirm the diagnosis. We may need to do a CAT scan of your head and an Influenzavirus antibody assay to rule out any other conditions. Additionally, we may need to evaluate you further and consider other respiratory therapy or physical therapy exercises to help you feel better.<SEP>The extracted entities are CAT scan of head (Head ct), Influenzavirus antibody assay, Physical therapy exercises; manipulation; and other procedures, Other respiratory therapy<EOS>
\n\n
Try to output:
### Instruction:\n'Learn to extract entities from the following medical questions.'\n\n### Input:\n
<CLS>{input}<SEP>The extracted entities are\n\n ### Output:
"""
\end{lstlisting}
\caption{The prompt template of Entity Recognition. The input is the question.}
\label{tab:ner}
\end{table*}

\begin{table*}[!thbp]
\centering
\begin{lstlisting}[language=Python, basicstyle = \ttfamily, breaklines = true]
template = """
    There are some knowledge graph path. They follow entity->relationship->entity format.
    \n\n
    {Path}
    \n\n
    Use the knowledge graph information. Try to convert them to natural language, respectively. Use single quotation marks for entity name and relation name. And name them as Path-based Evidence 1, Path-based Evidence 2,...\n\n


    Output:
    """


template = """
    There are some knowledge graph. They follow entity->relationship->entity list format.
    \n\n
    {neighbor}
    \n\n
    Use the knowledge graph information. Try to convert them to natural language, respectively. Use single quotation marks for entity name and relation name. And name them as Neighbor-based Evidence 1, Neighbor-based Evidence 2,...\n\n


    Output:
    """
\end{lstlisting}
\caption{The prompt templates of transfering path-based evidence subgraphs and neighbor-based evidence subgraphs to natural language.}
\label{tab:nlp}
\end{table*}

\begin{table*}[!thbp]
\centering
\begin{lstlisting}[language=Python, basicstyle = \ttfamily, breaklines = true]
 template = """
    You are an excellent AI doctor, and you can diagnose diseases and recommend medications based on the symptoms in the conversation.\n\n
    Patient input:\n
    {question}
    \n\n
    You have some medical knowledge information in the following:
    {instruction}
    \n\n
    What disease does the patient have? What tests should patient take to confirm the diagnosis? What recommened medications can cure the disease?
    """
\end{lstlisting}
\caption{The prompt templates of BM25 Retrieval and Embedding Retrieval. The input is the question and the most related document context.}
\label{tab:document}
\end{table*}

\begin{table*}[!thbp]
\centering
\begin{lstlisting}[language=Python, basicstyle = \ttfamily, breaklines = true]
def prompt_comparation(reference,output1,output2):
    template = """
    Reference: {reference}
    \n\n
    output1: {output1}
    \n\n
    output2: {output2}
    \n\n
    According to the facts of disease diagnosis and drug and tests recommendation in reference output, which output is better match. If the output1 is better match, output '1'. If the output2 is better match, output '0'. If they are same match, output '2'. 
    """
    prompt = template.format(reference=reference, output1=output1, output2=output2)
    response = openai.ChatCompletion.create(
      model="gpt-4",
      messages=[
          {"role": "system", "content": """You are an excellent AI doctor."""},
          {"role": "user", "content": prompt}
      ]
    ) 
    response_of_comparation = response.choices[0].message.content
    return response_of_comparation
\end{lstlisting}
\caption{The prompt template for GPT-4 rater to evaluate the factual correctness between our method and baselines, the reference is the answer or explanation label.}
\label{tab:gpt4}
\end{table*}

\end{appendix}

\end{document}